\documentclass[sigconf]{acmart}

\usepackage{multirow}

\newcommand\blfootnote[1]{%
  \begingroup
  \renewcommand\thefootnote{*}\footnotetext{#1}%
  \addtocounter{footnote}{-1}%
  \endgroup
}

\AtBeginDocument{%
  \providecommand\BibTeX{{%
    \normalfont B\kern-0.5em{\scshape i\kern-0.25em b}\kern-0.8em\TeX}}}

\copyrightyear{2021}
\acmYear{2021}
\setcopyright{acmcopyright}\acmConference[CIKM '21]{Proceedings of the 30th ACM International Conference on Information and Knowledge Management}{November 1--5, 2021}{Virtual Event, QLD, Australia}
\acmBooktitle{Proceedings of the 30th ACM International Conference on Information and Knowledge Management (CIKM '21), November 1--5, 2021, Virtual Event, QLD, Australia}
\acmPrice{15.00}
\acmDOI{10.1145/3459637.3482450}
\acmISBN{978-1-4503-8446-9/21/11}

\begin{document}
\fancyhead{}

\title{Match-Ignition: Plugging PageRank into Transformer \\ for Long-form Text Matching}



\author{Liang Pang$^1$, Yanyan Lan$^{2*}$, Xueqi Cheng$^3$}

\email{pangliang@ict.ac.cn, lanyanyan@tsinghua.edu.cn, cxq@ict.ac.cn}

\affiliation{%
   \institution{
   $^1$Data Intelligence System Research Center, \\
   Institute of Computing Technology, Chinese Academy of Sciences, Beijing, China \\
   $^2$Institute for AI Industry Research, Tsinghua University, Beijing, China \\
   $^3$CAS Key Lab of Network Data Science and Technology, \\
   Institute of Computing Technology, Chinese Academy of Sciences, Beijing, China \\
   }
}


\begin{abstract}\blfootnote{Corresponding author}
Neural text matching models have been widely used in community question answering, information retrieval, and dialogue.
However, these models designed for short texts cannot well address the long-form text matching problem, because there are many contexts in long-form texts can not be directly aligned with each other, and it is difficult for existing models to capture the key matching signals from such noisy data. 
Besides, these models are computationally expensive for simply use all textual data indiscriminately.   
To tackle the effectiveness and efficiency problem, we propose a novel hierarchical noise filtering model, namely Match-Ignition. The main idea is to plug the well-known PageRank algorithm into the Transformer, to identify and filter both sentence and word level noisy information in the matching process. 
Noisy sentences are usually easy to detect because previous work has shown that their similarity can be explicitly evaluated by the word overlapping, so we directly use PageRank to filter such information based on a sentence similarity graph. 
Unlike sentences, words rely on their contexts to express concrete meanings, so we propose to jointly learn the filtering and matching process, to well capture the critical word-level matching signals. Specifically, a word graph is first built based on the attention scores in each self-attention block of Transformer, and key words are then selected by applying PageRank on this graph. In this way, noisy words will be filtered out layer by layer in the matching process. 
Experimental results show that Match-Ignition outperforms both SOTA short text matching models and recent long-form text matching models. We also conduct detailed analysis to show that Match-Ignition efficiently captures important sentences and words, to facilitate the long-form text matching process.

\end{abstract}



\maketitle

{\fontsize{8pt}{8pt} \selectfont
\textbf{ACM Reference Format:}\\
Liang Pang, Yanyan Lan, Xueqi Cheng. 2021. Match-Ignition: Plugging PageRank into Transformer for Long-form Text Matching. In {\it Proceedings of the 30th ACM International Conference on Information and Knowledge Management (CIKM '21), November 1--5, 2021, Virtual Event, QLD, Australia.} ACM, New York, NY, USA, 10 pages. https://doi.org/10.1145/3459637.3482450}

\section{Introduction}
Semantic text matching is an essential problem in many natural language applications, such as community question answering~\cite{BiMPM}, information retrieval~\cite{DSSM}, and dialogue~\cite{lu2013deep}. 
Many deep text matching models have been proposed and gain good performance, such as representation based models~\cite{DSSM, CDSSM, RNN, MV-LSTM}, interaction based models~\cite{ARC-I, MP, SRNN, guo2016deep}, and their combinations~\cite{DUET, RE2, BERT}.

\begin{figure}
  \includegraphics[width=0.95\linewidth]{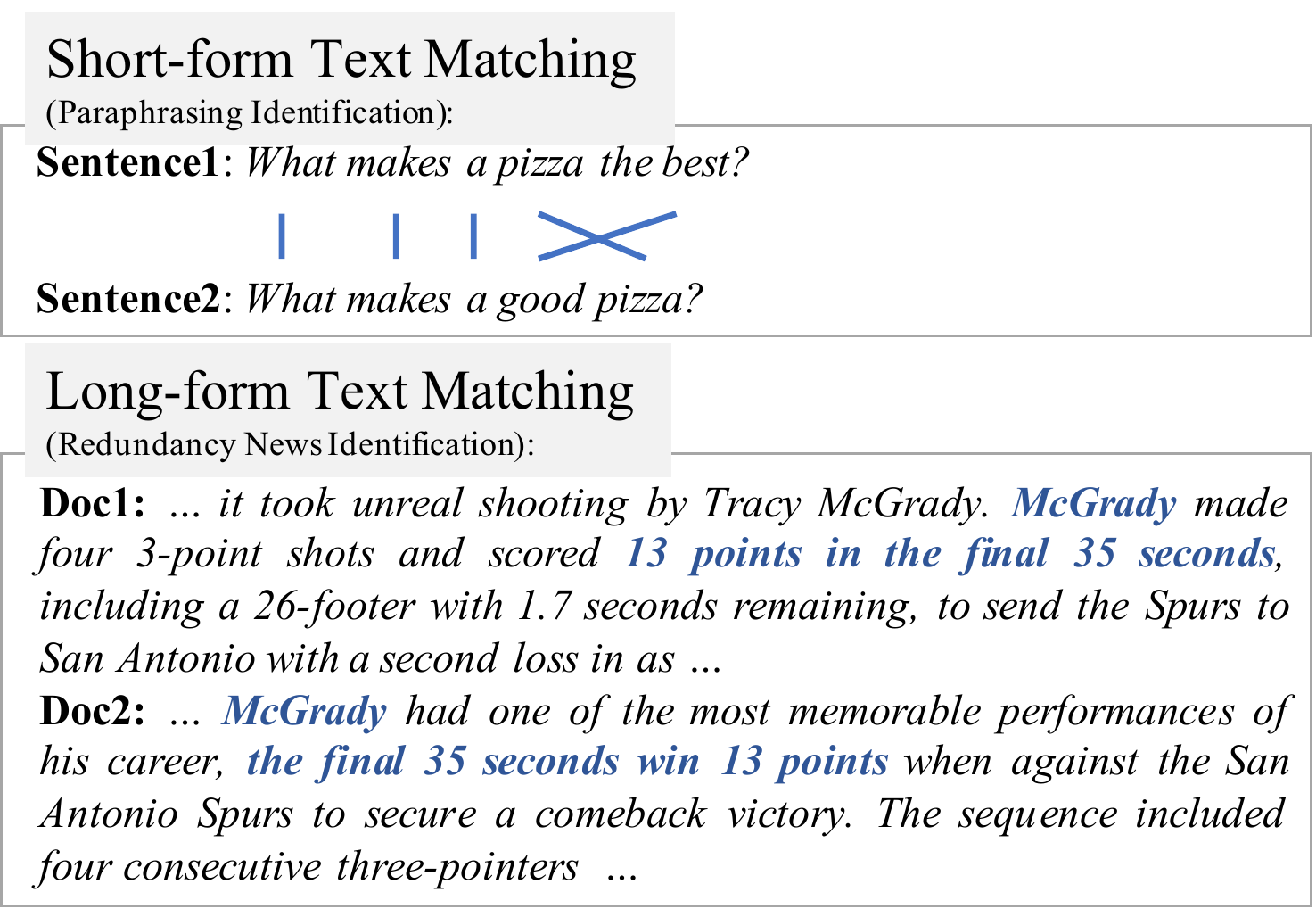}
  \caption{The top example is a short-form text matching for the paraphrasing identification, and the lines indicate the alignments between words from two sentences. The bottom example is a long-form text matching for the redundancy news identification, and the highlights indicate the important matching signals for the identity event of two news.}
  \label{fig:example}
\end{figure}

However, these models cannot be well applied to long-form text matching problems, which have attracted increasing attention in the field of news deduplication~\cite{liu-etal-2019-matching}, citation recommendation~\cite{yang2020512}, plagiarism detection~\cite{zhou-etal-2020-multilevel} and attachment suggestion~\cite{jiang2019semantic}. This is mainly because long-form text matching is quite different from the short-form text matching problem. For short-form text matching, almost every term in the short texts is critical to the matching score, because short text matching tasks are just like finding a reasonable semantic alignment between two sentences~\cite{MP}. For example, in paraphrasing identification, the major problem is to find the paraphrasing sentence for the given sentence. In this case, the matching score is mainly determined by the alignment between each word in the sentences, as shown in Figure~\ref{fig:example}. 

Long-form text matching has its own characteristics.
Firstly, long-form text matching cares more about the global semantic meanings rather than the bipartite alignment. The fine-grained matching signals between long-form texts are usually very sparse, which makes the existing short text matching models hard to figure out from huge noisy signals. For example, redundant news identification merely focuses on where/when the event happened and what the event is, instead of who posted this news and the detailed descriptions of the news. 
Secondly, long-form text intrinsically consists of a two-level structure, i.e.~sentences and words. Most existing short text matching approaches can only process text word by word while missing the sentence-level structure. For example, one sentence should be ignored entirely if it is irrelevant to the current document, e.g. advertisement, even though some of its internal words are relevant. 
Thirdly, long-form text matching contains a very long text by nature, which makes the existing short text matching models computational expensive because they have to treat every word indiscriminately and emphasize the sufficient interactions between words~\cite{RE2, BERT}. In practice, the long-form text often has to be truncated in the computation. For example, BERT only accepts text lengths of less than 512. These operations may hurt the final matching performance.
From these discussions, we can see that noise is the main challenge in long-form text matching, affecting both performance and efficiency.

In this paper, we propose a novel hierarchical noise filtering model, namely Match-Ignition, to distill the significant matching signals via the well-known link analysis algorithm PageRank~\cite{brin1998anatomy}. 
PageRank utilizes random walk on a graph to determine the importance of each node. In this way, the noises (i.e.~less important nodes) can be eliminated and the algorithm will be accelerated. Considering the two-level structures in the long-form text matching problem, our model contains two hierarchies, i.e.~sentence-level and word-level. In the sentence-level noise filtering process, the nodes are defined as sentences from a pair of long-form texts, and the link is defined as the similarities between each pair of sentences. That is to say, the similarities inside each long-form text and between the two long-form texts are both captured in our graph. Then the noisy sentences could be identified by PageRank score, and be directly removed. The word-level noise filtering process is jointly learned with the matching process. That is because each word relies on its context to express its concrete meanings, so noisy words need to be estimated dynamically during the matching process. To this end, we first apply the state-of-the-art Transformer to the texts, which well captures the contextual information among words. It turns out that the attention matrix in the self-attention block, the key component of Transformer, could be treated as a fully connected word-level similarity graph~\cite{guo2019star, zhao2019sparse, dai2019transformer}. PageRank is then applied to filter out noise words at each layer. We can see this technique is different from previous works~\cite{guo2019star, zhao2019sparse, dai2019transformer} which focus on eliminating links in the graph because our model focuses on filtering noisy words, i.e.,~nodes in the graph. Please note that attention weights could be directly applied to filter noisy words. The reason why we still use PageRank here is that the attention weights have been proven not reliable in explaining the importance of words in~\cite{brunner2019identifiability}. Furthermore, PageRank has the ability to consider the global importance of each word by value propagation on a graph, which is more thorough than attention weights.

We experiment on three long-form text matching tasks, news deduplication, citation recommendation, and plagiarism detection, including seven public datasets, e.g. CNSE, CNSS, AAN-Abs, AAN-Body, OC, S2ORC, and PAN. 
The experimental results show that Match-Ignition outperforms all baseline methods, including both short text matching models and recent long-form text matching models. The further detailed analysis demonstrates that Match-Ignition efficiently captures important matching signals in long-form text, which helps understand the matching process. Besides, we compare different noisy filtering methods to show the superiority of using PageRank.

\section{Related Work}
In this section, we first introduce the text matching models designed for short-form text matching, then review the most recent works for long-form text matching.

\textbf{Short-form Text Matching}
Existing text matching models fall into representation-based approaches, interaction-based approaches, and their combinations~\cite{guo2019deep}.

Representation-based matching approaches are inspired by the Siamese architecture~\cite{Siamese}. This kind of approach aims at encoding each input text in a pair into the high-level representations respectively based on a specific neural network encoder, and then the matching score is obtained by calculating the similarity between the two corresponding representation vectors. DSSM~\cite{DSSM}, C-DSSM~\cite{CDSSM}, ARC-I\cite{ARC-I}, RNN-LSTM~\cite{RNN} and MV-LSTM~\cite{MV-LSTM} belong to this category.
Interaction-based matching approaches are closer to the nature of the matching task to some extent since they aim at directly capturing the local matching patterns between two input text, rather than focusing on the text representations. The pioneering work includes ARC-II~\cite{ARC-I}, MatchPyramid~\cite{MP}, and Match-SRNN~\cite{SRNN}.
Recently, there has been a trend that the two aforementioned branches of matching models should complement each other, rather than being viewed separately as two different approaches. DUET~\cite{DUET} is composed of two separated modules, one in a representations-based way, and another in an interaction-based way, the final matching score is just their weighted-sum result. The attention mechanism is another way to combine the above two approaches, such as RE2~\cite{RE2} and BERT~\cite{BERT}.
However, these existing approaches for short-form text matching have limited success in long-form text matching settings, due to their inability to capture and distill the main information from long documents.
Besides, these models are computationally expensive because they simply use all textual data indiscriminately in the matching process.

\textbf{Long-form Text Matching}
Due to the lack of the public datasets and the efficient algorithms, few work directly focuses on the long-form text matching and further explores the application scenarios of it.
In recent years, since the pioneering work SMASH proposed by \citeauthor{jiang2019semantic}~\cite{jiang2019semantic}, they are the first to point out that long-form text matching (e.g. source text and target text both are long-form text) has a wide range of application scenarios, such as attachment suggestion, article recommendation, and citation recommendation. They propose a hierarchical recurrent neural network under Siamese architecture which is a kind of representation-based matching approach. It synthesizes information from different document structure levels, including paragraphs, sentences, and words.
SMITH model~\cite{yang2020512} follows the SMASH's settings, then utilizes powerful pre-trained language model BERT~\cite{BERT} as their key component and breaks the 512 tokens limitation to build a representation-based matching approach.
Instead of using BERT as a component, TransformerXL~\cite{dai2019transformer} and Longformer~\cite{beltagy2020longformer} try to directly extent the Transformer structure towards the long-form text by introducing a sliding window or memory strategy into it. However, they are designed for general natural language understanding, not for text matching tasks.
Another work on long-form text matching is Concept Interaction Graph (CIG)~\cite{liu-etal-2019-matching}, which concerns modeling the relation between two documents, e.g. same event or story. It can be treated as an interaction-based matching approach, which selects a pair of sentences based on their concepts and similarities. Besides, they also construct two types of duplicate news detection datasets, which are labeled by professional editors.

All the previous works ignore the fact that long-form text provides overabundant information for matching, that is to say, there are usually many noises in the setting of long-form text matching.
This phenomenon also is discussed in query-document matching tasks~\cite{guo2016deep, pang2017deeprank, hui2017pacrr, fan2018modeling}, which is a short to long text matching because a query is a short-form text and a document is a long-form text.
DeepRank~\cite{pang2017deeprank} treats query and document differently, in their model, each query term acts as a filter that picks out text spans in the document which contain this query term. That is to say, query irrelevant text spans are the noise that can be ignored in the matching process.
PACRR~\cite{hui2017pacrr} also has similar findings, they filter document words using two kinds of processes, 1) keep first $k$ terms in the document or 2) retain only the text that is highly relevant to the given query.
These previous works provide strong evidence that our noise filtering motivation can be effective for long-form text matching problems.

\begin{figure*}
  \includegraphics[width=0.9\linewidth]{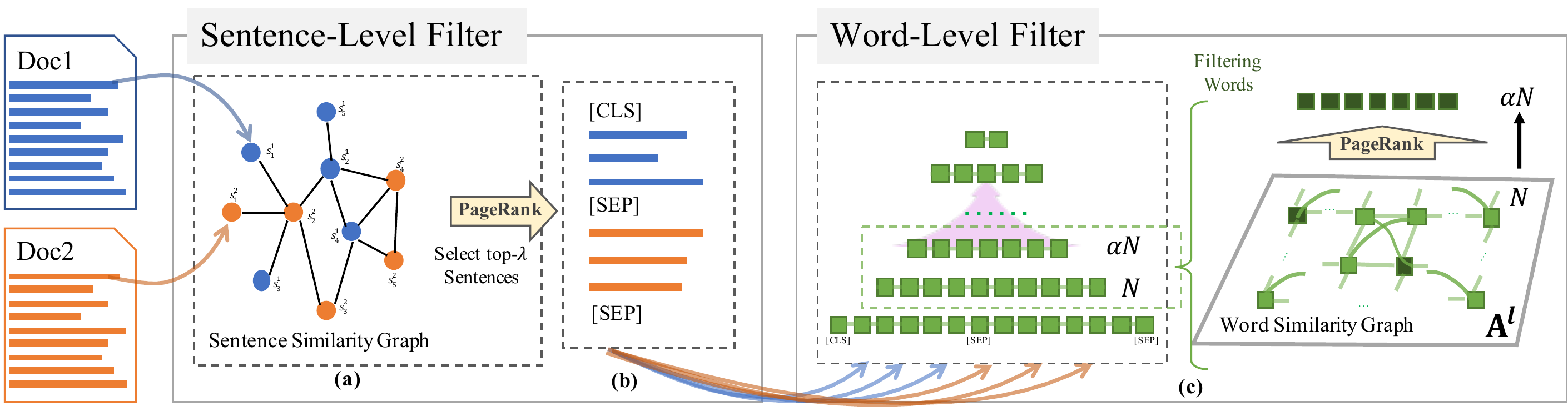}
  \caption{The overall architecture of Match-Ignition. (a) represents the sentence-level filter, (b) represents the outputs of the sentence-level filter, and (c) represents the word-level filter.}\label{fig:model}
\end{figure*}

\section{Match-Ignition}

In this section, we first introduce the two components of Match-Ignition. They are sentence-level noise filter and word-level noise filter, shown in Figure~\ref{fig:model}(a) and Figure~\ref{fig:model}(c) respectively.
After that, the model training details are described in the last subsection.

\subsection{Sentence-level Noise Filtering}

To enable the application of graph-based ranking algorithms PageRank to natural languages, such as documents, a graph is needed to build that represents the relation between sentences. 
TextRank~\cite{mihalcea2004textrank} makes it possible to form a sentence extraction algorithm, which can identify key sentences in a given document. It becomes a mature approach in automatic summarization.
The most direct way is to apply the TextRank algorithm on each long-form text independently, so that we can reduce the length of each long-form text by summarizing. 
However, the goal of long-form text matching is to find the matching signals between a pair of texts, which is different from  summarization task that extracts key information from one text. Therefore, directly applying the TextRank algorithm to each text independently leads to the problem of loss of matching signals.
 
Inspired by the previous works~\cite{pang2017deeprank, hui2017pacrr}, they tell us that two texts can help each other for noise detection, so that both long-form texts should be represented in one graph to involve the matching information across two texts.
Firstly, sentences in both long-form texts are collected together to form a united sentence collection.
Formally, two long-form texts are first split into sentences, denoted as $d_s = [s^1_1, s^1_2, \dots, s^1_{L_1}]$ and $d_t = [s^2_1, s^2_2, \dots, s^2_{L_2}]$, where $L_1$ and $L_2$ are the number of sentences in $d_s$ and $d_t$ respectively. 
The united sentence collection $\mathcal{S} = \{s^1_1, s^1_2, \dots, s^1_{L_1}, s^2_1, s^2_2, \dots, s^2_{L_2}\}$ then have $L_1+L_2$ elements. 
Thus, the sentence similarity graph can be constructed by evaluating the sentence pair similarities in the united sentence collection $\mathcal{S}$. 
The sentence similarity is defined as the same as in TextRank~\cite{mihalcea2004textrank}, to measures the overlapping word ratio between two sentences:
\begin{equation}
	Sim(s_i, s_j) = \frac{|\{w_k|w_k \in s_i, w_k \in s_j\}|}{\log(|s_i|)+\log(|s_j|)},\;\; s_i, s_j \in \mathcal{S},
\end{equation}
where $w_k$ denotes the word in the sentence, $|\cdot|$ denotes the length of the sentence or word set, and $s_i, s_j$ are two sentences in the united sentence collection $\mathcal{S}$.
To make sentence similarity sparsity e.g. returns 0 most of the time, we remove the stopwords in the sentences before we calculate the similarities. Thus, the final sentence similarity graph has sparse links.
Finally, a PageRank algorithm is applied to this constructed sentence similarity graph, to get the important score of each sentence. To balance the information coming from different long-form texts for the following step, the top $\lambda$ sentences are extracted for each long-form texts respectively. Thus, both texts contain $\lambda$ sentences as their digestion, which we called a sentence-level filter.

As shown in Figure~\ref{fig:model}(b), the selected sentences are concatenated as a text sequence, which starts with [CLS] token and separates by [SEP] token. It is then treated as the input of the model in the word-level filter.
Note that the hyper-parameter $\lambda$ should be neither too small to lose a lot of information, nor too large to make text extremely long. A suitable $\lambda$ can yield a moderate text sequence, which length is just less than the BERT max input length.

PageRank algorithm can also be used at the word level if we can define a word-by-word relation graph. However, sentences are adjectives from each other, noise in this level is discrete than an entire sentence can be removed in an unsupervised way, while a word relies on its context to express concrete meanings, noise in this level is continuous that should be estimated during the model training.
Therefore, we need to construct a graph within the Transformer model structures.

\subsection{Word-level Noise Filtering}
To filter the noise in the word level, a word-level graph needs to be constructed first in the inherent transformer structure (Sec~\ref{sec:transformer_as_graph}). After that, the traditional PageRank algorithm is required to implement as a tensor version, for better to embed into the transformer structure (Sec~\ref{sec:tensor_pagerank}). Finally, we propose our plug PageRank to the Transformer model for word-level noise filtering (Sec~\ref{sec:pagerank-transformer}).

\subsubsection{Transformer as a Graph} \label{sec:transformer_as_graph}

Transformer architecture~\cite{vaswani2017attention} boosts the natural language processing a lot, where most well-known models are a member of this family, such as BERT~\cite{devlin-etal-2019-bert}, RoBERTa~\cite{liu2019roberta}, and GPT2~\cite{radford2019language}. They achieve state-of-the-art performance in almost all NLP tasks, e.g. named entity recognition, text classification, machine translation, and also text semantic matching. For long-form text matching, we also adopt this architecture.

The self-attention block is the main component in Transformer architecture, which figure out how important all the other words in the sentence are for the contextual word around it. Thus, the self-attention block builds the relations between words, which can be viewed as a fully connected graph among words~\cite{guo2019star, zhao2019sparse, dai2019transformer}.
Knowing that the updated word representations are simply the sum of linear transformations of representations across all the words, weighted by their importance.
It makes full use of the attention mechanism in deep neural networks to update word representations. As have shown in~\cite{vaswani2017attention}, the attention function can be formalized as a scaled dot-product attention with inputs $\mathbf{H}^{l}$: 

\begin{equation}\label{eq:attention}
	\begin{aligned}
		\mathbf{H}^{l+1} & = \mathrm{Attn}(\mathbf{Q}^{l}, \mathbf{K}^{l}, \mathbf{V}^{l}) 	
		 = \mathrm{Softmax}\left(\frac{\mathbf{Q}^{l}(\mathbf{K}^{l})^T}{\sqrt{E}}\right) \mathbf{V}^{l}
		= \mathbf{A}^{l} \mathbf{V}^{l}, \\
	\end{aligned}
\end{equation}
where $\mathbf{Q}^{l} = \mathbf{W_Q}^{l}\mathbf{H}^{l} \in \mathbb{R}^{N \times E}$ denote the attention query matrices, $\mathbf{K}^{l} = \mathbf{W_K}^{l}\mathbf{H}^{l} \in \mathbb{R}^{N \times E}$ the key matrix, and $\mathbf{V}^{l} = \mathbf{W_V}^{l}\mathbf{H}^{l} \in \mathbb{R}^{N \times E}$ the value matrix. $N$ denotes the number of words in a text, and $E$ denotes the dimensions of the representation. The attention mechanism can be explained as: for each attention query vector in $\mathbf{Q}$, it first computes the dot products of the attention query with all keys, aiming to evaluate the similarity between the attention query and each key. Then, it is divided each by $\sqrt{E}$, and applies a softmax function to obtain the weights on the values, denotes as $\mathbf{A}^{l}$. Finally, the new representation of the attention query vector is calculated as weighed sum of values.
Getting this dot-product attention mechanism to work proves to be tricky bad random initializations can de-stabilize the learning process. It can be overcome by performing multiple `heads' of attention and concatenating the result:
\begin{equation} \label{eq:adj_mat}
	\begin{aligned}
		&\mathbf{H}^{l+1} = \mathrm{Concat}(head_1, \cdots, head_H) \mathbf{O}^{l}, \\
		&head_k = \mathrm{Attention}(\mathbf{Q}^{kl}, \mathbf{K}^{kl}, \mathbf{V}^{kl}) = \mathbf{A}^{kl} \mathbf{V}^{kl},
	\end{aligned}
\end{equation}
where $\mathbf{Q}^{kl}$, $\mathbf{K}^{kl}$ and $\mathbf{V}^{kl}$ are of the $k$-th attention head at layer $l$ with different learnable weights, $\mathbf{O}^{l}$ down-projection to match the dimensions across layers, $H$ is the number of the heads in each layer and $L$ is the number of the layers.

If we treat each word as a node in a graph, they update their representations by aggregating all other contextual word representations, just like messages passing from other neighbor nodes in graph neural network~\cite{scarselli2008graph}.
Thus, for self-attention block, it can be treated as a fully-connected word graph, where its adjacency matrix is the transpose of the word-by-word similarity matrix $\mathbf{A}^{kl}$.

\subsubsection{PageRank in A Tensor View}
	\label{sec:tensor_pagerank}
PageRank~\cite{brin1998anatomy} is a graph-based ranking algorithm, which is essentially a way of deciding the importance of a vertex within a graph, and recursively attracts global information from the entire graph.
Formally, given a graph $G(V, E)$, where $V=\{v_1, v_2, \dots, v_N\}$ is a set of nodes and $E$ is the links between these nodes. The goal is to determine the order of these nodes that the more important node has a higher rank. The PageRank value on each node $v_i$, denotes as $u_i$, is used to indicate the importance of the node $v_i$.
For convenience, we define $\mathbf{A}$ as the adjacency matrix, that $\mathbf{A}_{ij}$ denotes the $v_i$ has a link from $v_j$ with weight $\mathbf{A}_{ij}$. $\mathbf{A}$ is also a stochastic matrix because each column sums up to 1. 
At the initial step all $u_i$ have the same value $1/N$, denotes that all nodes are equally important. At each following step , then PageRank value $u_i$ is updated using other nodes and links pointed to it,
\begin{equation}
	u_i = \sum\nolimits_{v_j \in V} \mathbf{A}_{ij} \cdot u_j.
\end{equation}
After several iterations, the PageRank values $u_i$ will converge to a set of stable values $u_i$, and that is the solution of PageRank.

To implement PageRank in a tensor-based computational framework, such as TensorFlow~\cite{abadi2016tensorflow} or PyTorch~\cite{paszke2019pytorch}, we need a tensor version of PageRank algorithm.
Let $\mathbf{u}^t=[u^t_1, u^t_2, \dots, u^t_n]$ to be a vector of length $N$, that obtains all nodes PageRank values at step $t$. Then, PageRank can be rewritten as, 
\begin{equation} \label{eq:pagerank}
	\mathbf{u}^{t+1} = \mathbf{A} \mathbf{u}^t.
\end{equation}
To solve the problem of isolated nodes, a stable version of PageRank is proposed~\cite{brin1998anatomy} and adopted by our work, 
\begin{equation}
	\mathbf{u}^{t+1} = d \mathbf{A} \mathbf{u}^t + (1-d)/N \cdot \mathbb{I}, 
\end{equation}
where $d \in [0, 1]$ is a real value to determine the ratio of the two parts, and $\mathbb{I}$ is a vector of length $N$ with all its values are 1.  The factor $d$ is usually set to 0.85, and this is the value we are also using in our implementation.

In practice, the number of the iteration step $T$ is set to a fixed value for computational efficiency. Thus, $\mathbf{u}^t$ is the final PageRank scores for each $v_i \in V$, and the larger of PageRank denotes the more importance of this node in the current graph, thus we can filter out the nodes with small PageRank values.

\subsubsection{Plug PageRank in Transformer}
	\label{sec:pagerank-transformer}
	
In this section, we propose a novel approach that plugs PageRank in the Transformer model to filter the noise at the word level. Notice that, word-level noise is composite, thus need to be estimated dynamically during the matching process. In each self-attention block, an inherent PageRank algorithm is utilized to dynamically filter the noisy words, which can reduce the sequence length layer by layer.	

Standard Transformer structure, which has been selected as our base model structure, has $L$ layers of multi-head self-attention blocks, stacked one after another, and maintains the same sequence length $N$ at each layer.
From the description in Section~\ref{sec:transformer_as_graph}, we have known that self-attention block in Transformer can be treated as a word-by-word graph, which can be specified using an adjacency matrix $(\mathbf{A}^{kl})^\top$ at $k$-th head and $l$-th layer in Eq~\ref{eq:adj_mat}.
The word-level noise filtering process is once per layer, thus we need to average the effects of all adjacency matrices across different heads in the $l$-th layer,
\begin{equation}
	\mathbf{A}^{l} = \frac{1}{H}\sum\nolimits_{k=1}^{H}\mathbf{A}^{kl}.
\end{equation}

Because $\mathbf{A}^{l}$ is the output of row-wise Softmax function, each row of $\mathbf{A}^{l}$ sum to 1. Thus, $(\mathbf{A}^{l})^\top$ is a stochastic matrix, which can be treated as the adjacency matrix in a graph.
With above observation, we substitute $(\mathbf{A}^{l})^\top$ into Eq~\ref{eq:pagerank} and yield:
\begin{equation}
	\mathbf{u}^{t+1} = d (\mathbf{A}^{l})^\top \mathbf{u}^t + (1-d)/N \cdot \mathbb{I}. 
\end{equation}

Iteratively solving the equation above, we then get the PageRank values for all words/nodes in the $(l-1)$-th layer, denote as $\mathbf{u}$. 
Thus, $\mathbf{u}$ represents the importance of the words in the $(l-1)$-th layer.
After applying the attention mechanism to the words in the $(l-1)$-th layer, we get a list of new word representations as to the input of $l$-th layer.
To filter noisy words, we have to estimate the importance of the words/nodes in $l$-th layer, which can be evaluated by redistributing the importance of the word in $(l-1)$-th layer under the distribution $\mathbf{A}^{l}$, thus the word importance scores are $\mathbf{r} = \mathbf{A}^{l} \mathbf{u}$.
Finally, we can reduce the sequence length at $l$-th layer by removing the nodes which have the small values in $\mathbf{r}$.

In this work, we design a strategy that remove the percentage $\alpha \in [0\%, 100\%]$ nodes per layer, so that the $l$-th layer has $(\alpha)^{l-1} \cdot N$ nodes. The hyper-parameter $\alpha$ is called a word reduction ratio.
For example, let $L=12, N=400$, if we set $\alpha$ to 10\%, the numbers of nodes at each layer are 400, 360, 324, 291, 262, 236, 212, 191, 172, 154, 139, 125.

For the BERT model, some words are too special to be removed, such as [CLS] token and [SEP] token. If the model occasionally removes these tokens during the training, it will lead to an unstable training process. It also affects the overall performance. 
Therefore, a token mask is designed to keep these tokens across all the layers.

\textbf{Discussions}: 
Many previous works~\cite{guo2019star, zhao2019sparse, dai2019transformer} have also noticed the relation between Transformer and graph.
Star-Transformer~\cite{guo2019star} adds a hub node to model the long-distance dependence and eliminates the links far from 3-term steps.
TransformerXL~\cite{dai2019transformer} uses a segment-level recurrence with a state reuse strategy to remove all the links between words in different segments, so that can break the fixed-length limitation.
Sparse-Transformer~\cite{zhao2019sparse} explicitly eliminate links in which attention scores are lower than the threshold to make the attention matrix sparse.
All of these previous works focus on eliminating links in the graph, while in this work, we focus on filtering noise words, as well as nodes, in the graph.

\subsection{Model Training}
The sentence-level filter is the heuristic approach that does not need a training process.
Thus, in this section, we only consider model training for the word-level filter component.

For the model training of word-level filter, we adopt the ``pre-training + fine-tuning'' paradigm as in BERT. 
In this paradigm, the pre-trained Transformer is firstly obtained using a large unlabeled plain text in an unsupervised learning fashion. Then, the Transformer plugging PageRank at each layer is fine-tuned using the supervised downstream task.
Note that word-level filters do not change the parameters in the original Transformer, due to all the parameters in the Transformer are input sequence length independent. Therefore, change the sequence length layer by layer does not affect the structure of the Transformer. Benefit from the good property of PageRank-Transformer, we can directly adopt a publicly released Transformer model, such as BERT or RoBERTa trained on a large corpus, as our pre-trained model.

In the fine-tuning step, we add the PageRank module in each self-attention layer, without introducing any additional parameters. The objective function for long-form text matching task is a binary cross-entropy loss:
\begin{equation}
	\mathcal{L} = - \sum\nolimits_i y_i \log{p_i} + (1-y_i) \log(1-p_i), 
\end{equation}
where $p_i$ is the probability represents the matching score, generated by the representation of [CLS], and $y_i$ is the ground-truth label.

\section{Experiments}
In this section, we conduct experiments and in-depth analysis on three long-form text matching tasks to demonstrate the effectiveness and efficiency of our proposed model.

\begin{table}
	\caption{Description of evaluation datasets, AvgWPerD denotes average number of words per document and AvgSPerD denotes the average number of sentences per document.}
	\label{Table:dataset}
	\centering
	\scalebox{0.95}{
		\begin{tabular}{cccccc}
			\toprule
			Dataset & AvgWPerD & AvgSPerD & Train & Dev & Test\\
			\midrule
			CNSE & 982.7 & 20.1 & 17,438 & 5,813 & 5,812 \\
			CNSS & 996.6 & 20.4 & 20,102 & 6,701 & 6,700 \\
			\hline
			AAN-Abs & 122.7 & 4.9 & 106,592 & 13,324 & 13,324 \\
			AAN-Body & 3270.1 & 111.6 & 104,371 & 12,818 & 12,696 \\
			OC & 190.4 & 7.0 & 240,000 & 30,000 & 30,000 \\
			S2ORC & 263.7 & 9.3 & 152,000 & 19,000 & 19,000 \\
			\hline
			PAN & 1569.7 & 47.4 & 17,968 & 2,908 & 2,906 \\
			\bottomrule
		\end{tabular}
	}
\end{table}

\subsection{Datasets}

We use seven public datasets in our experiments, and their detailed statistics are shown in Table~\ref{Table:dataset}.

\textbf{News Deduplication}: 
For this task, we use two datasets, i.e.~Chinese News Same Event dataset (CNSE) and the Chinese News Same Story dataset (CNSS) released in~\cite{liu-etal-2019-matching}. They are constructed based on large Chinese news articles collected from major Internet news, which cover diverse topics in the open domain~\footnote{Datasets are available at \url{https://github.com/BangLiu/ArticlePairMatching}}. CNSE is designed to identify whether a pair of news articles report the same breaking news (or event), and CNSS is used to identify whether they belong to the same series of news stories, labeled by professional editors. The negative samples in the two datasets are not randomly generated. Document pairs that contain similar keywords are selected and exclude samples with TF-IDF similarity below a certain threshold. Finally, we follow the settings in~\cite{liu-etal-2019-matching} and split either dataset into training, development, and testing set with the portion of instances 6:2:2. 

\textbf{Citation Recommendation}: 
Citation recommendations can help researchers find related works and finish paper writing more efficiently. Given the content of a research paper and a candidate citation paper, the task aims to predict whether the candidate should be cited by the paper.
In our experiment, four datasets are used for this task, i.e.~AAN-Abs, AAN-Body, OC, and S2ORC. 
AAN-Abs and AAN-Body are both constructed from the AAN dataset~\cite{radev2013acl}, which contains computational linguistics papers published on ACL Anthology from 2001 to 2014, along with their metadata. For the AAN-Abs dataset released in~\cite{zhou-etal-2020-multilevel}, each paper's abstract and its citations' abstracts are extracted and treated as positive pairs, and negative instances are sampled from uncited papers. For the AAN-Body dataset, we follow the same setting described in the previous work ~\cite{jiang2019semantic,yang2020512,tay2020long}, where they remove the reference sections to prevent the leakage of ground-truth and remove the abstract sections to increase the difficulty of the task. Besides, we also use the same training, development, and testing splitting released in~\cite{tay2020long}. The OC dataset~\cite{bhagavatula2018content} contains about 7.1M papers major in computer science and neuroscience. 
The S2ORC dataset~\cite{lo2020s2orc} is a large contextual citation graph of 8.1M open access papers across broad domains of science. The papers in S2ORC are divided into sections and linked by citation edges. 
AAN-Abs, OC, and S2ORC are pre-processed, split, and released by~\cite{zhou-etal-2020-multilevel}, for a fair comparison, we adopt the same settings in our experiments.

\textbf{Plagiarism Detection}:
The detection of plagiarism has received considerable attention to protect the copyright of publications. It is a typical long-form text matching problem because even partial text reuse will identify plagiarism between two documents. For this task, we use the PAN dataset~\cite{potthast2013overview}, which collects web documents with various plagiarism phenomena. Human annotations are employed to indicate the text segments that are relevant to the plagiarism both in the source and suspicious documents. Follow the settings of~\cite{yang-etal-2016-hierarchical}, the positive pairs are constructed by the segment of the source document and the suspicious document annotated as plagiarism, while the negative pairs are subsequently constructed by replacing the source segment in the positive pair with a segment from the corresponding source documents which is not annotated as being plagiarised. 
Note that the aforementioned AAN-Abs, OC, S2ORC, and PAN datasets can be directly downloaded~\footnote{https://xuhuizhou.github.io/Multilevel-Text-Alignment/}.

\textbf{Evaluation Metrics}: Since all the above tasks are binary classification, we use accuracy and F1 to act as the evaluation measures, similar to~\cite{liu-etal-2019-matching, zhou-etal-2020-multilevel}. Specifically for each method, we perform training for 10 epochs and then choose the epoch with the best validation performance to evaluate on the test set.

\subsection{Baselines and Experimental Settings}
We adopt three types of baseline methods for comparison, including traditional term-based retrieval methods, deep learning methods for short-form text matching, and recent deep learning methods for long-form text matching.

We select three traditional term-based methods for comparison, i.e.~BM25~\cite{robertson2009probabilistic}, LDA~\cite{blei2003latent} and SimNet~\cite{liu-etal-2019-matching}. BM25 is one of the most popular traditional term-based methods. The experimental results on CNSE and CNSS datasets are directly brought from \cite{liu-etal-2019-matching}, while others are implemented by ourselves. LDA is a famous topic model, which is used here to demote the matching method by computing similarities between the two texts represented by topic modeling vectors. We do not implement it by ourselves, and the results are directly from \cite{liu-etal-2019-matching}.
SimNet first extracts five text-pair similarities and conducts classification by a multi-layer neural network, whose results are brought from \cite{liu-etal-2019-matching}.
	
Considering deep learning methods for short-form text matching, we compare four types of models, including representation-based models, i.e. DSSM~\cite{DSSM}, C-DSSM~\cite{CDSSM}, and ARC-I~\cite{ARC-I}; interaction-based models, i.e. ARC-II~\cite{ARC-I} and MatchPyramid~\cite{MP}; hybrid models, i.e. DUET~\cite{DUET} and RE2~\cite{RE2}, and the pretraining matching model BERT-Finetuning~\cite{devlin-etal-2019-bert}. The results of DSSM, C-DSSM, ARC-I, ARC-II, MatchPyramid, and DUET on CNSE and CNSS dataset, are directly borrowed from the previous work~\cite{liu-etal-2019-matching}, which uses the implementations from MatchZoo~\cite{fan2017matchzoo}~\footnote{\url{https://github.com/NTMC-Community/MatchZoo}}.
RE2~\cite{RE2} is implemented using released code by the author~\footnote{\url{https://github.com/alibaba-edu/simple-effective-text-matching}} with the default configuration, e.g. 300-dimensions pre-trained word vectors provided by {\em Glove.840B} and 30-epochs training.
BERT-Finetuning~\cite{devlin-etal-2019-bert} is fine-tuned on text matching tasks based on a large-scale pre-training language model, e.g.~BERT for Chinese `\emph{bert-base-chinese}' and BERT for English `\emph{bert-base-uncased}' in the Transformers library~\footnote{\url{https://github.com/huggingface/transformers}}. For each of the pretraining models, we finetune it 10 epochs on the training set.

For the methods specially designed for the long-form text matching problem, we first focus on traditional hierarchical models, e.g. HAN~\cite{yang-etal-2016-hierarchical} and its variance GRU-HAN~\cite{zhou-etal-2020-multilevel}, GRU-HAN-CDA~\cite{zhou-etal-2020-multilevel} and SMASH~\cite{jiang2018beyond}. 
Then, we compare our model with a group of BERT-based hierarchical models, e.g. MatchBERT~\cite{zhou-etal-2020-multilevel}, BERT-HAN~\cite{zhou-etal-2020-multilevel}, BERT-HAN-CDA~\cite{zhou-etal-2020-multilevel} and SMITH~\cite{yang2020512}.
Finally, we consider some matching models by representing each text by a pretrained model specifically designed for the long-form text, like TransformerXL~\cite{dai2019transformer} and Longformer~\cite{beltagy2020longformer}.
The results of GRU-HAN, GRU-HAN-CDA, BERT-HAN, and BERT-HAN-CDA on AAN-Abs, OC, S2ORC, and PAN datasets are from the previous work~\cite{zhou-etal-2020-multilevel}.
The results of HAN, SMASH, MatchBERT and SMITH on the AAN-Body dataset are from the previous work~\cite{yang2020512}.
TransformerXL-Finetuning and Longformer-Finetuning are implemented using the Transformers library. The pretrained model we selected are `\emph{transfo-xl-wt103}' for TranformerXL and `\emph{allenai/longformer-base-4096}' for Longformer. For each of the pre-trained models, we finetune it 10 epochs on the training set.

Specially, for CNSE and CNSS datasets, Concept Interaction Graph (CIG) model~\cite{liu-etal-2019-matching} is the state-of-the-art approach, which generates the representation for each vertex, and then uses a GCN to obtain the matching score. So we compare with three representative models of this approach, i.e.~CIG-Siam-GCN, CIG-Sim\&Siam-GCN, and CIG-Sim\&Siam-GCN-Sim$^g$. The results are obtained by implementations based on their released code~\footnote{\url{https://github.com/BangLiu/ArticlePairMatching}}.

The hyper-parameters of our Match-Ignition model are listed below. For the sentence-level filter, the number of selected sentences per text $\lambda$ is set to 5, the $d$ in PageRank algorithm defined in Eq~\ref{eq:pagerank} is set to 0.85. For the word-level filter, we adopt a pre-trained BERT model for Chinese, e.g. `bert-base-chinese', which contains 12 heads and 12 layers. The words filtering ratio $\alpha$ is set to 10\%, that is to say, we remove 10\% words per layer. The fine-tuning optimizer is Adam~\cite{kingma2014adam} with the learning rate $10^{-5}$. $\beta_1=0.9, \beta_2=0.999, \epsilon=10^{-8}$, and batch size is set to 8.
The model is built based on the Transformers library using PyTorch~\cite{paszke2019pytorch}. The source code will be released at \url{https://github.com/pl8787/Match-Ignition}.

\begin{table}
	\caption{Experimental results on the news deduplication task, e.g. CNSE and CNSS datasets. Significant performance degradation with respect to Match-Ignition is denoted as (-) with $\mathbf{p}$-value $\leq$ 0.05. We only do significant test on the models reimplemented from the source code, while the results bring from \cite{liu-etal-2019-matching} do not test due to the lack of the detailed predictions.}
	\label{Table:exp_news}
	\centering
	\small
	\scalebox{0.92}{
	\begin{tabular}{clllll}
		& & \multicolumn{2}{c}{CNSE Dataset} & \multicolumn{2}{c}{CNSS Dataset} \\
		\toprule
		& Model & Acc & F1 & Acc & F1\\
		\midrule
		\multirow{3}*{I} & BM25~\cite{robertson2009probabilistic} & 69.63 & 66.60 & 67.77 & 70.40 \\
		& LDA~\cite{blei2003latent} & 63.81 & 62.44 & 67.77 & 70.40 \\
		& SimNet~\cite{liu-etal-2019-matching} & 71.05 & 69.26 & 70.78 & 74.50 \\
		\midrule
		\multirow{8}*{II} & ARC-I~\cite{ARC-I} & 53.84 & 48.68 & 50.10 & 66.58 \\
		& ARC-II~\cite{ARC-I} & 54.37 & 36.77 & 52.00 & 53.83 \\
		& DSSM~\cite{DSSM} & 58.08 & 64.68 & 61.09 & 70.58 \\
		& C-DSSM~\cite{CDSSM} & 60.17 & 48.57 & 52.96 & 56.75 \\
		& MatchPyramid~\cite{MP} & 66.36 & 54.01 & 54.01 & 62.52 \\
		& DUET~\cite{DUET} & 55.63 & 51.94 & 52.33 & 60.67 \\
		& RE2~\cite{RE2} & 80.59$^-$ & 78.27$^-$ & 84.84$^-$ & 85.28$^-$ \\
		& BERT-Finetuning~\cite{BERT} & 81.30$^-$ & 79.20$^-$ & 86.64$^-$ & 87.08$^-$ \\
		\midrule
		\multirow{3}*{III} & CIG-Siam-GCN~\cite{liu-etal-2019-matching} & 74.58$^-$ & 73.69$^-$ & 78.91$^-$ & 80.72$^-$ \\
		& CIG-Sim\&Siam-GCN~\cite{liu-etal-2019-matching} & 84.64$^-$ & 82.75$^-$ & 89.77$^-$ & 90.07$^-$ \\
		& CIG-Sim\&Siam-GCN-Sim$^g$~\cite{liu-etal-2019-matching} & 84.21$^-$ & 82.46$^-$ & 90.03$^-$ & 90.29$^-$ \\
		\midrule
		\multirow{1}*{IV} & Match-Ignition & \textbf{86.32} & \textbf{84.55} & \textbf{91.28} & \textbf{91.39} \\ 
		\bottomrule
	\end{tabular}
	}
\end{table}

\begin{table*}[t]
    \caption{Experimental results on the citation recommendation task, e.g. AAN-Abs, OC, and S2ORC datasets and the plagiarism detection task, e.g. PAN dataset. Significant performance degradation with respect to Match-Ignition is denoted as (-) with $\mathbf{p}$-value $\leq$ 0.05. We only do significant test on the models reimplemented from the source code, while the results bring from \cite{zhou-etal-2020-multilevel} do not test due to the lack of the detailed predictions.}\label{Table:exp_cite}
    \small
    \centering
    \scalebox{0.95}{
    \begin{tabular}{clllllll|ll}
        & & \multicolumn{2}{c}{AAN-Abs Dataset} & \multicolumn{2}{c}{OC Dataset} & \multicolumn{2}{c}{S2ORC Dataset} & \multicolumn{2}{c}{PAN Dataset} \\
        \toprule
        & Model & Acc & F1 & Acc & F1 & Acc & F1 & Acc & F1\\
        \midrule
        \multirow{1}*{I} & BM25~\cite{robertson2009probabilistic} & 67.60$^-$ & 68.00$^-$ & 80.32$^-$ & 80.38$^-$ & 76.47$^-$ & 76.53$^-$ & 61.59$^-$ & 62.47$^-$\\
        \midrule
        
        \multirow{2}*{II} & RE2~\cite{RE2} & 87.81$^-$ & 88.04$^-$ & 94.53$^-$ & 94.57$^-$ & 95.27$^-$ & 95.34$^-$ & 61.97$^-$ & 58.30$^-$\\
        & BERT-Finetune~\cite{BERT} & 88.10$^-$ & 88.02$^-$ & 94.87$^-$ & 94.87$^-$ & 96.32$^-$ & 96.29$^-$ & 59.11$^-$  & 69.66$^-$ \\
        \midrule
        
        \multirow{4}*{III} & GRU-HAN~\cite{zhou-etal-2020-multilevel} & 68.01 & 67.23 & 84.46 & 82.26 & 82.36 & 83.28 & 75.70 & 75.88 \\
        & GRU-HAN-CDA~\cite{zhou-etal-2020-multilevel} & 75.08 & 75.18 & 89.79 & 89.92 & 91.59 & 91.61 & 75.77 & 76.71 \\      
        & BERT-HAN~\cite{zhou-etal-2020-multilevel} & 73.36 & 73.51 & 86.31 & 86.81 & 90.67 & 90.76 & 87.57 & 87.36 \\
        & BERT-HAN-CDA~\cite{zhou-etal-2020-multilevel} & 82.03 & 82.08 & 90.60 & 90.81 & 91.92 & 92.07 & 86.23 & 86.19 \\
        \midrule
        
        \multirow{2}*{IV} & TransformerXL-Finetune~\cite{dai2019transformer} & 83.85$^-$ & 83.24$^-$ & 91.61$^-$ & 91.79$^-$ & 92.50$^-$ & 92.39$^-$ & 58.25$^-$ & 69.07$^-$ \\
        & Longformer-Finetune~\cite{beltagy2020longformer} & 88.06$^-$ & 88.41$^-$ & 94.76$^-$ & 94.74$^-$ & 96.31$^-$ & 96.29$^-$ & 56.61$^-$ & 69.74$^-$ \\
        \midrule
        
        \multirow{1}*{V} & Match-Ignition & \textbf{89.62} & \textbf{89.64} & \textbf{95.70} & \textbf{95.71} &  \textbf{96.97} & \textbf{96.97} & \textbf{89.37} & \textbf{89.42} \\
        \bottomrule
    \end{tabular}
    }
\end{table*}

\subsection{Experimental Results} 
The performance comparison results of Match-Ignition against baseline models are shown in Table~\ref{Table:exp_news} and Table~\ref{Table:exp_cite}. From these experimental results, we can obtain the following summaries:

1) Comparing Match-Ignition with existing deep learning models for short-form text matching, we can see that Match-Ignition outperforms all types of the existing short-form text matching models on both CNSE and CNSS datasets. Specially, it performs significantly better than two strong baselines, e.g. the hybrid model RE2 and the pretraining model BERT-Finetuning. For the tasks of citation recommendation and plagiarism, we compare the Match-Ignition model with the two strongest short-form text matching methods in the news deduplication task. The results in Table~\ref{Table:exp_cite} II show that Match-Ignition also significantly outperforms RE2 and BERT-Finetuning, which demonstrates the superiority of Match-Ignition against existing short-form text matching models.

2) Then we Compare Match-Ignition with the current state-of-the-art methods in the three tasks, including graph-based method CIG, hierarchical methods HAN and its variants.
For the news deduplication task, the Match-Ignition model significantly outperforms the state-of-the-art method CIG-Sim\&Siam-GCN-Sim$^g$, as shown in Table~\ref{Table:exp_news} III. That is because CIG is usually affected by noisy concept terms, while our model has the ability to filter noisy information in the learning process. For the other two tasks, as we can see in Table~\ref{Table:exp_cite} III, Match-Ignition outperforms the state-of-the-art hierarchical methods HAN and HAN-CDA. Note that another newly proposed state-of-the-arts model SMITH-WP+SP is also an extension of the hierarchical method to tackle long-form text matching problems. However, they do not use the AAN-Abs dataset but conduct their experiments on the context-only version of the AAN dataset, namely AAN-Body. To compare with it, we apply our model to AAN-Body dataset. As we can see from the results in Table~\ref{Table:exp_ann} II, the Match-Ignition model also outperforms the SMITH-WP+SP model.

3) Comparing the Match-Ignition model with the matching models based on pretrained long-form text representation models, e.g. TransformerXL and Longformer. The experimental results in Table~\ref{Table:exp_cite} IV show that comparing with the finetuned version of TransformerXL and Longformer, the Match-Ignition model achieves better performances. Especially in the plagiarism detection task, noises affect the performances a lot, since the goal of TransformerXL and Longformer is to preserve the information in the long-form text as much as possible. Note that TransformerXL and Longformer only release their English versions, which are not applicable in the Chinese dataset, thus we do not list their results for the news deduplication task on CNSE and CNSS datasets.

Furthermore, we compare with another branch of the state-of-the-art hierarchical methods, e.g. SMASH and SMITH, on the AAN-Body dataset in Talbe~\ref{Table:exp_ann}. We do not implement them on other datasets because 1) SMASH does not provide code for model construction and training; 2) SMITH needs to pretrain on large-scale data and then finetune. So we implement our model on the dataset they used, e.g. the AAN-Body dataset, to achieve a fair comparison. The experimental results show that Match-Ignition also outperforms these baseline methods.

\begin{table}[t]
    \caption{Experimental results on AAN-Body dataset make a fair comparison for our Match-Ignition model and the long-form text matching models proposed in \cite{yang2020512}.}\label{Table:exp_ann}
    \small
    \centering
    \begin{tabular}{clll}
        & & \multicolumn{2}{c}{AAN-Body Dataset} \\
		\toprule
		& Model & Acc & F1\\
		\midrule
		\multirow{1}*{I} & BM25~\cite{robertson2009probabilistic} & 59.66 & 59.90 \\
		\midrule
		\multirow{1}*{II} & RE2~\cite{RE2} & 80.15 & 79.25 \\
		\midrule
		\multirow{4}*{III} & HAN~\cite{yang-etal-2016-hierarchical} & 82.19 & 82.57 \\
		& SMASH~\cite{jiang2019semantic} & 83.75 & 82.78 \\
		& MatchBERT~\cite{yang2020512} & 83.55 & 82.93 \\
        & SMITH-WP+SP~\cite{yang2020512} & 85.36 & 85.43 \\
        \midrule
        \multirow{1}*{IV} & Match-Ignition & \textbf{89.92} & \textbf{89.91} \\
        \bottomrule
    \end{tabular}
\end{table}

\begin{table}
	\caption{Ablation study of the two-level noise filtering mechanisms in Match-Ignition.}
	\label{Table:ablation}
	\centering
	\small
	\begin{tabular}{lcccc}
		\toprule
		& \multicolumn{2}{c}{CNSE} & \multicolumn{2}{c}{CNSS} \\
		Model & Acc & F1  & Acc & F1 \\
		\midrule
		Match-Ignition & 86.32 & 84.55 & 91.28 & 91.39 \\
		\; $\cdot$ Sentense-level Filter Only & 84.11 & 82.17 & 91.04 & 91.07 \\
		\; $\cdot$ Word-level Filter Only & 80.31 & 76.39 & 91.10 & 91.18 \\
		BERT-Finetune & 81.30 & 79.20 & 86.64 & 87.08 \\
		\bottomrule
	\end{tabular}
\end{table}

\begin{table}
	\caption{The impact of word reduction ratio $\alpha$ and the execution time of these models.}
	\label{Table:ana_word_denoise}
	\centering
	\small
	\begin{tabular}{lcccccc}
		\toprule
		Words Reduc- & \multicolumn{2}{c}{CNSE} & \multicolumn{2}{c}{CNSS} & \multicolumn{2}{c}{Time per batch}\\
		tion Ratio $\alpha$ & Acc & F1  & Acc & F1 & Train & Eval\\
		\midrule
		0\% & 84.11 & 82.17 & 91.04 & 91.07 & 1.73s & 0.42s \\
		5\% & 85.68 & 83.65 & 90.70 & 90.73 & 1.58s & 0.37s\\
		10\% & \textbf{86.32} & \textbf{84.55} & \textbf{91.28} & \textbf{91.39} & 1.33s & 0.31s\\
		20\% & 82.55 & 79.66 & 90.25 & 90.21 & 1.07s & 0.21s\\
		\bottomrule
	\end{tabular}
\end{table}

\begin{table}
	\caption{Comparison results with other word-level noise filtering strategies.}
	\label{Table:word_filter}
	\centering
	\small
	\begin{tabular}{lcccc}
		\toprule
		 & \multicolumn{2}{c}{CNSE} & \multicolumn{2}{c}{CNSS} \\
		Word-level Filter & Acc & F1  & Acc & F1 \\
		\midrule
		Random           & 80.38 & 79.19 & 87.68 & 88.20 \\
		Embedding Norm   & 80.54 & 78.15 & 84.92 & 85.03 \\
		Attention Weight & 85.52 & 83.34 & 89.91 & 89.88 \\
		PageRank         & 86.32 & 84.55 & 91.28 & 91.39 \\
		\bottomrule
	\end{tabular}
\end{table}

\subsection{Ablation Study} \label{sec:ablation}
Now we conduct an ablation study to investigate the two-level noise filtering strategies in Match-Ignition on news deduplication.

\subsubsection{Investigations on the sentence-level filtering}

In Table~\ref{Table:ablation}, the `Sentence-level Filter Only' and `Word-level Filter Only' model denotes the result by only using the sentence-level and word-level filter, respectively. Therefore, the sentence-level filter is critical on both CNSE and CNSS datasets: 1) without it, the accuracy on CNSE will degrade from 86.32\%/91.28\% to 80.31\%/91.10 on CNSE and CNSS, respectively; 2) Match-Ignition with sentence-level filter only outperforms the strong baseline `BERT-Finetune' by 3.5\% and 5.1\% w.r.t. accuracy on CNSE and CNSS, respectively. 

\subsection{Case Study}
\begin{figure}
  \includegraphics[width=1\linewidth]{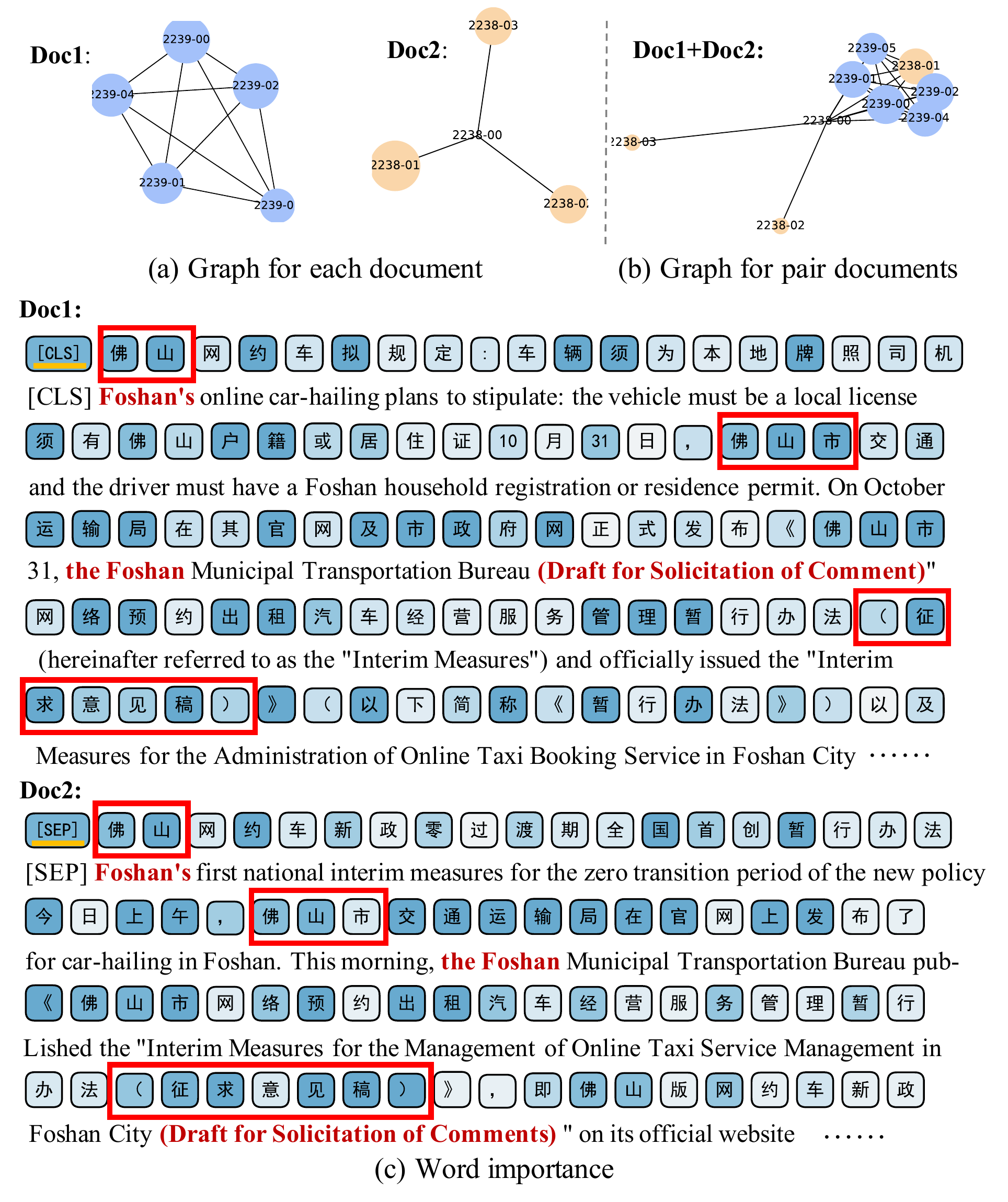}
  \caption{(a) sentence graph for each document using TextRank, (b) sentence graph built in Match-Ignition, each sentence is a node in the graph, its color represents the document it belongs to and its size represents the importance (PageRank value). (c) illustrates the word importances, and the darker color means the more important word.}\label{fig:case}
\end{figure}

\subsubsection{Investigations on word-level filtering}

Still, from Table~\ref{Table:ablation}, we can see the effect of the word-level filtering: 1) the performance increases 2.6\% and 0.3\% from `Sentence-level Filter Only' model to the full version Match-Ignition model on CNSE and CNSS, respectively; 2) though Match-Ignition with word-level filter only cannot beat BERT-Finetune on CNSE, it outperforms BERT-Finetune by 5.1\% on CNSS. Therefore, word-level filtering also plays an important role in Match-Ignition. 

Then we study the impact of word reduction ration $\alpha$, which is a major hyper-parameter in the word-level filter because it determines how many words/nodes should be deleted in each layer.
Specifically, we evaluate four words reduction ratio, where $\alpha=0\%$ means the word-level filter is turned off. From the results shown in Table~\ref{Table:ana_word_denoise}, we can see that too small or large a value of $\alpha$ leads to bad performance, and $\alpha=10\%$ yields the best performances on both CNSE and CNSS datasets. 

We also demonstrate the efficiency of the Match-Ignition model with different $\alpha$ as in Table~\ref{Table:ana_word_denoise}. Please note that the sentence-level filter executes very fast, comparing to evaluating Transformer. So the efficiency on sentence-level filter can be ignored. Theoretically, the major time cost in Match-Ignition is computing the word-by-word similarity matrices in self attention blocks in the Transformer model. Let $N$ denotes the text length and $L$ denotes the number of layers, the computation cost can be approximated by
	${\text{TimeCost}}(\alpha) \approx \sum\nolimits_{l=0}^{L-1} (1-\alpha)^{2l}$
where ${\text{TimeCost}}(0\%)\approx12$ and ${\text{TimeCost}}(20\%)\approx2.76$, thus $\alpha=20\%$ is 4 times faster than $\alpha=0\%$ in theory. In our experiments, we use a single 12G Nvidia K80 GPU with batch size 8. The efficiency results in Table~\ref{Table:ana_word_denoise} show that $\alpha=20\%$ is 1.6 times faster than $\alpha=0\%$ at the training stage and 2 times faster at the evaluation stage. 

Furthermore, we compare different types of word-level filtering strategies. {\em Random} stands for the method which randomly selects words at each layer, and {\em Embedding Norm} selects words depending on its embedding norms. {\em Attention Weight} uses the attention weight to determine the importance of the word, which has been proven to be a special case of {\em PageRank}, i.e.~without propagation on the graph. From the results in Table~\ref{Table:word_filter}, PageRank achieves the best results, demonstrating the importance of word selecting strategies in word-level filtering.

To illustrate the Match-Ignition model more intuitively, we give an example from the CNSE dataset to visualize the sentence-level graph (Fig~\ref{fig:case}~(a)(b)) and word importance (Fig~\ref{fig:case}~(c)). 

Specifically, Figure~\ref{fig:case}~(a) demonstrates the graph by directly applying TextRank on each document separately, and Figure~\ref{fig:case}~(b) shows the constructed sentence-level graph built-in Match-Ignition. The difference indicates the rationality of our model. For example, sentence `2238-01' are equally important as `2238-02' and `2238-03' in Figure~\ref{fig:case}~(a). While it becomes much more important in Figure~\ref{fig:case}~(b) because it has more connections with the sentences in Doc1. Therefore, our model is capable to capture the key sentences in the matching process, by considering connections both inside and between two documents.

Fig~\ref{fig:case}~(c) show the word importance in different colors, where darker color indicates a higher importance score. Specifically, the importance score is computed based on the number of layers retaining the word, which shows the importance of each word in the whole matching process. The results are accordant with human understanding. For example, the location `the Foshan' and the name of the policy `Draft for Solicitation of Comment' is important for determining the matching degree of the news, which indeed obtain a higher importance score in the model, 
as highlighted with rectangles. Furthermore, the results show that special tokens like [CLS] and [SEP] are also important for long-form text matching. That is because [CLS] token acts as the global information aggregator and [SEP] token acts as a separator of the two texts, which are two crucial indicators in long-form text matching.

\section{Conclusion}
In this paper, we propose a novel hierarchical noise filtering approach for the long-form text matching problem. The novelty lies in the employment of the well-known PageRank algorithm to identify and filter both sentence-level and word-level noisy information, which can be viewed as a generalized version of using attention weight with propagation on the graph. We conduct extensive experiments on three typical long-form text matching tasks including seven public datasets, and the results show that our proposed model significantly outperforms both short-form text matching models and recent state-of-the-arts long-form text matching models. 

In the future, we plan to investigate how to jointly learn the sentence-level and word-level noise filter in Match-Ignition. In addition, we would like to study the relation between Match-Ignition and graph neural network, and whether there exists a graph neural network-based model to achieve the two-level noise filtering in long-form text matching.

\begin{acks}
This work was supported by National Natural Science Foundation of China (NSFC) under Grants No. 61906180, No. 61773362 and No. 91746301, National Key R\&D Program of China under Grants 2020AAA0105200.  
\end{acks}

\balance
\bibliographystyle{ACM-Reference-Format}
\bibliography{match}
\end{document}